\documentclass[conference]{IEEEtran}
\usepackage{times}
\usepackage{amsfonts}
\usepackage{amsmath}
\usepackage{bm}
\newtheorem{example}{\hspace*{-1em}\textbf{Example}}

\usepackage[numbers]{natbib}
\usepackage{multicol}
\usepackage[bookmarks=true]{hyperref}
\usepackage{graphicx}
\usepackage{color}
\usepackage[it,small]{caption}
\usepackage{subcaption}
\usepackage{dsfont}
\usepackage{wrapfig}
\newcommand{\argmax}{\operatorname{arg\,max}}

\newcommand{\todo}[1]{\textcolor{blue}{\textbf{#1}}}
\newcommand{\robobrain}{RoboBrain}

 \belowdisplayskip \abovedisplayskip
\renewcommand{\baselinestretch}{0.97}

\newlength\savedwidth

\newlength{\sectionReduceTop}
\newlength{\sectionReduceBot}
\newlength{\subsectionReduceTop}
\newlength{\subsectionReduceBot}
\newlength{\abstractReduceTop}
\newlength{\abstractReduceBot}
\newlength{\captionReduceTop}
\newlength{\captionReduceBot}
\newlength{\subsubsectionReduceTop}
\newlength{\subsubsectionReduceBot}
\newlength{\headerReduceTop}
\newlength{\figureReduceBot}

\newlength{\horSkip}
\newlength{\verSkip}

\newlength{\equationReduceTop}

\newlength{\figureHeight}
\begin{document}

% paper title
\title{RoboBrain: \\Large-Scale Knowledge Engine for Robots}

% You will get a Paper-ID when submitting a pdf file to the conference system
% \author{Author Names Omitted for Anonymous Review. Paper-ID 100}

\author{Ashutosh Saxena, Ashesh Jain, Ozan Sener, Aditya Jami, Dipendra K Misra, Hema S Koppula\\
Department of Computer Science, Cornell University and Stanford University.\\
Email: \{asaxena, ashesh, ozansener, adityaj, dipendra, hema\}@cs.stanford.edu}

\maketitle
\maketitle

% !TEX root = robobrain.tex

\begin{abstract}
In this paper we introduce a knowledge engine, which learns and shares knowledge representations, for robots to carry out a variety of tasks. Building such an engine brings with it the challenge of dealing with multiple data modalities including symbols, natural language, haptic senses, robot trajectories, visual features and many others. The \textit{knowledge} stored in the engine comes from multiple sources including physical interactions that robots have while performing tasks (perception, planning and control), knowledge bases from the Internet and learned representations from several robotics research groups. 

We discuss various technical aspects and associated challenges such as modeling the correctness of knowledge, inferring latent information and formulating different robotic tasks as  queries to the knowledge engine. We describe the system architecture and how it supports different mechanisms for users and robots to interact with the engine. Finally, we demonstrate its use in three important research areas: grounding natural language, perception, and planning, which are the key building blocks for many robotic tasks. This knowledge engine is a collaborative effort and we call it  \robobrain{}.
\vskip .1in
\end{abstract}


\IEEEpeerreviewmaketitle

% !TEX root = robobrain.tex
% \vskip .1in
\section{Introduction}
\begin{figure}
 \centering
\includegraphics[width=\linewidth]{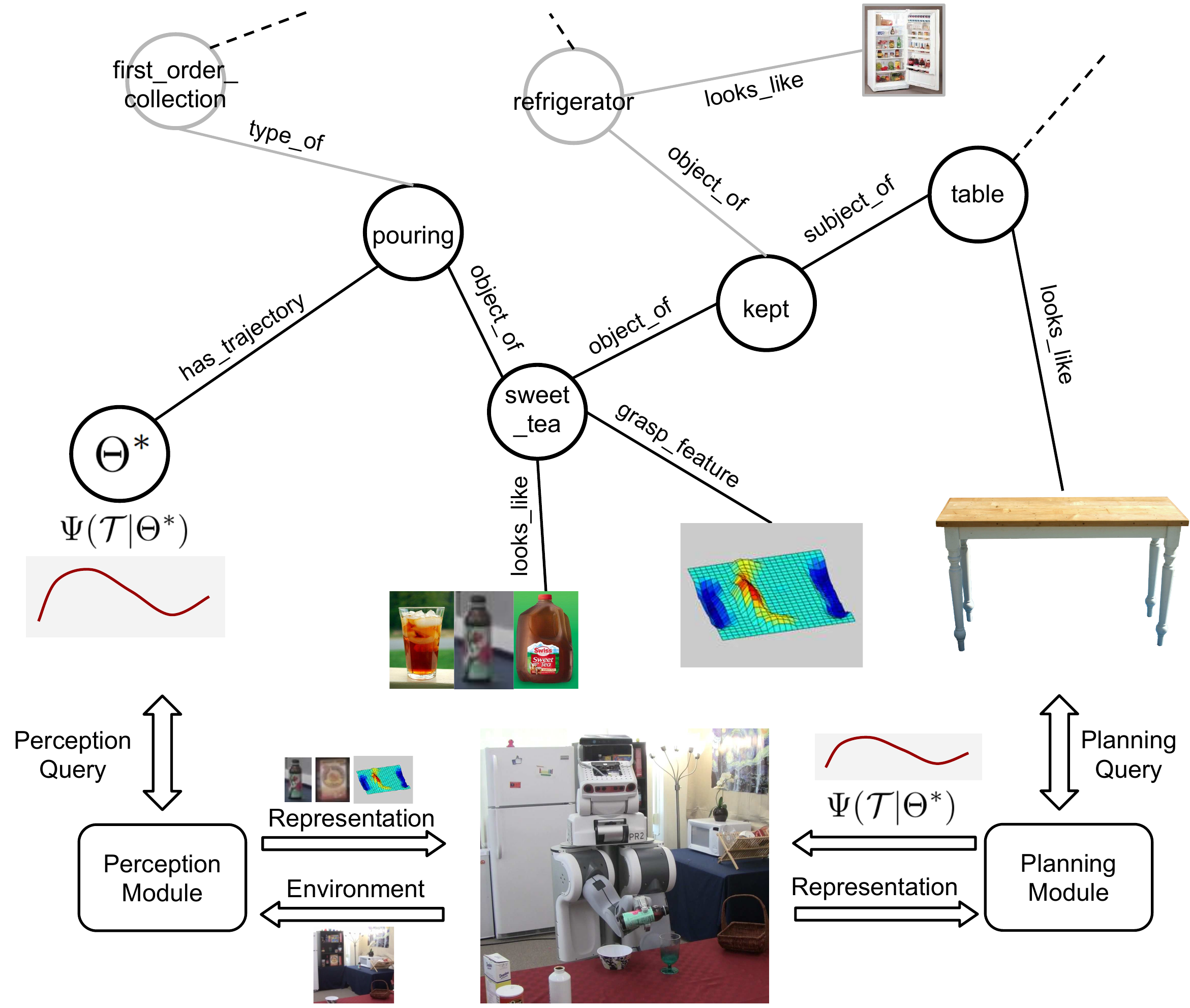}
 \caption{\textbf{An example showing a robot using \robobrain{} for performing tasks.}  The robot is asked ``Bring me sweet tea from the kitchen", where
 it needs to translate the instruction into the perceived state of the environment.
\robobrain{} provides useful knowledge to the robot for performing the task:
 (a) sweet tea can be kept on a table or inside a refrigerator,
 (b) bottle can be grasped in certain ways,
 (c) opened sweet tea bottle needs to be kept upright,
 (d) the pouring trajectory should obey user preferences of moving slowly to pour, and
 so on.
 }
 \label{fig:intro}
  \vskip -.2in
 \end{figure}

Over the last decade, we have seen many successful applications of large-scale knowledge systems.
%  many inclusive web services have been developed  by successfully combining information at a large-scale.
Examples include  Google knowledge graph \cite{dong2014knowledge}, IBM Watson~\cite{ferrucci2010}, Wikipedia,
 and many others.
These systems know answers to many of our day-to-day questions, and
 not crafted for a specific task, which makes them valuable for humans.
Inspired by them, researchers have aggregated domain specific knowledge by mining
data~\cite{dbpedia2007, freebase2008}, and processing natural
language~\cite{nell2010}, images~\cite{imagenet2009} and
speech~\citep{mohamed2011deep}.  These sources of knowledge are specifically designed for humans, and their  human centric design makes them of limited use for robots---for example, imagine a robot querying a search engine
 for how to ``bring sweet tea from the kitchen" (Figure~\ref{fig:intro}).

 %Contrary to humans, for whom incomplete and ambiguous instructions may suffice,
 In order to perform a task, robots require access to a large variety of information with finer details for performing
 perception, planning, control and natural language understanding. When asked to bring sweet tea, as shown in Figure~\ref{fig:intro}, the robot
 would need access to the knowledge for grounding the language symbols into physical entities,
  the knowledge that sweet tea can either be on a table or in a fridge, and the knowledge for inferring the
 appropriate plans for grasping and manipulating objects. Efficiently handling this joint knowledge representation across different
 tasks and modalities is still an open problem.

In this paper we present \robobrain{} that allows robots to learn and share such representations of knowledge.
We learn these knowledge representations from a variety of sources, including interactions that robots have while performing perception,
planning and control, as well as natural language and visual data from the Internet.
Our representation considers several modalities including symbols, natural language, visual or shape features, haptic properties, and so on. \robobrain{} connects this knowledge from various sources and allow robots to perform diverse tasks by jointly reasoning over multiple data modalities.  %Moreover, our experiments suggest that by connecting different modalities and projects, robots can share knowledge and collectively improve their abilities.

\begin{figure*}[t]
\centering
\includegraphics[width=.95\linewidth, height=2.6in, clip=true, trim=0 40 0 0]{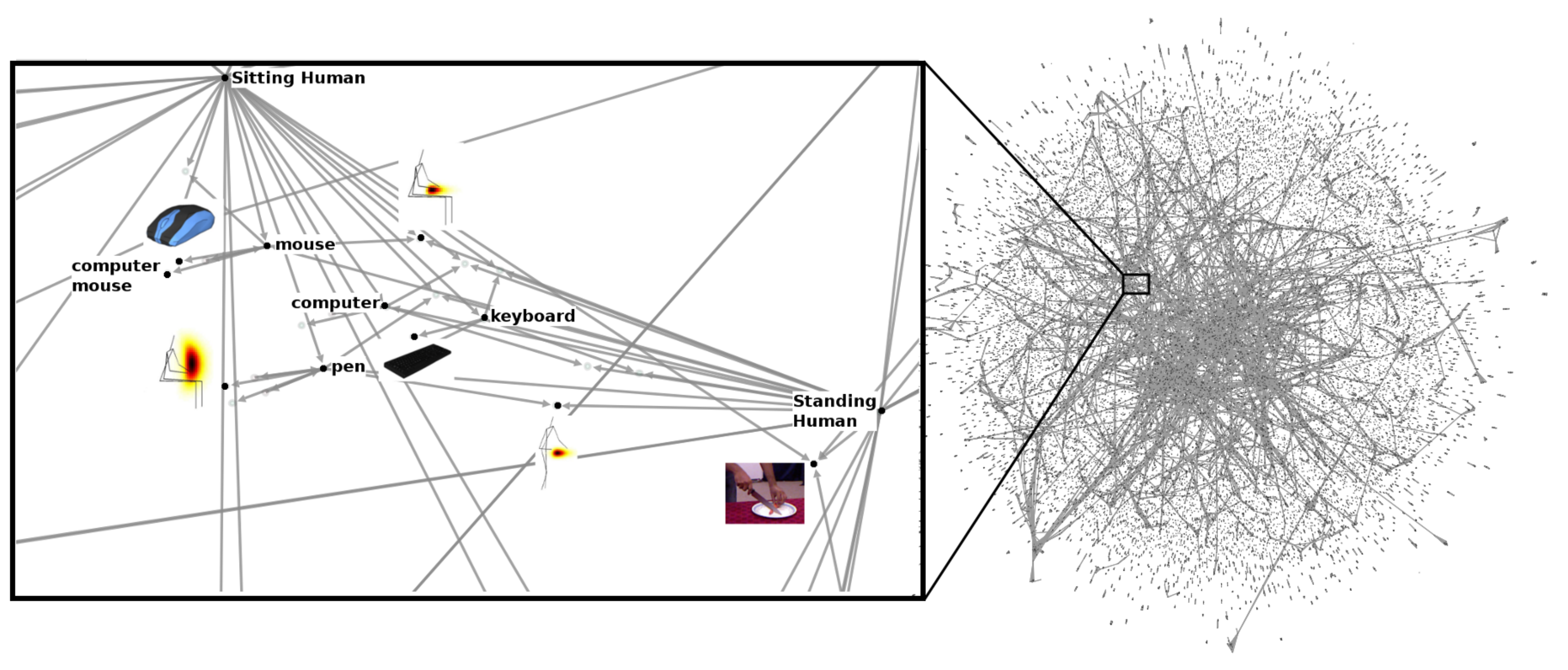}
\vskip -.05in
\caption{\textbf{A visualization of the \robobrain{} graph} on Nov 2014, showing about 45K nodes and 100K directed edges.
%  from the RoboBrain's knowledge graph.
The left inset shows a zoomed-in view of a small region of the graph with rendered media. This illustrates the relations between multiple modalities namely images, heatmaps, words and human poses. \textit{For high-definition graph visualization, see: \url{https://sites.google.com/site/robotknowledgeengine/}}}
\label{fig:graph}
\vskip -.2in
\end{figure*}

\robobrain{} enables sharing from multiple sources by representing the knowledge in a graph structure.
% RKE enables knowledge sharing by representing the knowledge in a graph representation.
% Thcontaining
% out of the
% information that robots need for performing various tasks.
%Knowledge in  RKE comes from multiple sources, which is integrated into the graph structure of RKE.
Traversals on the \robobrain{} graph allow robots to gather the specific information they need for a task. This includes the  semantic information, such as different grasps of the same object, as well as the functional knowledge, such as spatial constraints (e.g., a bottle is kept on the table and not the other way around).
The key challenge lies in building this graph from a variety of knowledge sources while ensuring dense connectivity across nodes. Furthermore, there are several  challenges in building a system that allows concurrent and distributed update, and retrieval operations.

We  present use of \robobrain{} on three robotics applications in the area of grounding natural language, perception and planning.
For each application we show usage of \robobrain{} \textit{as-a-service}, which allow researchers to effortlessly use the state-of-the-art algorithms. We also present experiments to show that sharing knowledge representations through \robobrain{} improves existing language grounding and path planning algorithms.

\iffalse
We present \robobrain{}-as-a-service scenario which lets robotics researchers to easily integrate available state of the art algorithms and collected structured multi-modal data,
sharing semantic knowledge from unstructured internet sources scenario, and fusion of multiple algorithms scenario. Our experimental evaluation suggests that \robobrain{} allows effortless development of complicated robotics applications.
They also suggest that \robobrain{} can replace tedious manual labelling process with simple queries over unstructured internet sources. It can also automatically choose an algorithm based on the inferred beliefs.
\fi

\robobrain{} is a collaborative project that we support by designing a large-scale cloud architecture.
%and we designed a large-scaled cloud architecture to enable this collaboration efficiently.
In the current state, \robobrain{} stores and shares knowledge across several research projects~\cite{tellex2011understanding,misra2014tell,jainsaxena2013_trajectorypreferences,jiang-hallucinatinghumans-labeling3dscenes-cvpr2013,jiang2012humancontext,KoppulaRSS2013,wulenzsaxena2014_hierarchicalrgbdlabeling,lenz2013_deeplearning_roboticgrasp} and Internet knowledge sources~\citep{cyc1995,wordnet1998}.
% RKE has knowledge from \cite{misra2014tell,tellex2011understanding,jainsaxena2013_trajectorypreferences,jiang-hallucinatinghumans-labeling3dscenes-cvpr2013}.
We believe as more research projects contribute knowledge to \robobrain{}, it will not only improve the concerned project but will
also be beneficial for the robotics community at large.
%their robots perform better but we also believe this will be beneficial for the robotics community at large.
%The RKE is under open Creative Commons Attribution license\footnote{http://creativecommons.org/licenses/by/2.5, Extra conditions may apply depending on the data sources, see anonymous.me for detailed information.} and is
%avalable at: \texttt{\url{http://anonymous}}

This is our first paper introducing \robobrain{}. It summarizes the key ideas and challenges in building a knowledge engine for robots. The goal of the paper is to present an overall view of the \robobrain{}, its architecture, functionalities, and demonstrate its application to robotics.
\iffalse
The goal of this paper is to summarize the key ideas in building \robobrain{}, present
the architecture details, and demonstrate its applications to robotics.  \todo{It is
our first paper introducing \robobrain{}, and is written as a systems paper.}
\fi
 In
Section~\ref{sec:graph} we formally define the \robobrain{} graph and describe its
system architecture in Section~\ref{sec:system}. In order for robots to use
\robobrain{} we propose the Robot Query Library in Section~\ref{sec:raquel}. In
Section~\ref{sec:applications} we present different robotic applications using \robobrain{}.

% !TEX root = robobrain.tex
\vspace*{2\sectionReduceTop}
\section{Related Work\label{sec:relatedwork}}
\vspace*{2\sectionReduceBot}
We now describe some works related to \robobrain{}. We first give an overview of the existing knowledge bases and describe how \robobrain{} differs from them. We then describe some works in robotics that can benefit from \robobrain{}, and also discuss some of the related on-going efforts.

\smallskip

% !TEX root = robobrain.tex
\noindent
\textbf{Knowledge bases.}
Collecting and representing a large amount of information in a knowledge base (KB) has been widely studied in the
areas of data mining, natural language processing and machine learning. Early seminal works have manually created KBs for the study of common sense knowledge (Cyc \cite{cyc1995}) and lexical knowledge (WordNet \cite{wordnet1998}). With the growth of Wikipedia, KBs started to use crowd-sourcing (DBPedia \cite{dbpedia2007}, Freebase \cite{freebase2008}) and automatic information extraction (Yago \cite{yago2007,yago22013}, NELL \cite{nell2010}) for mining knowledge.

One of the limitations of these KBs is their strong dependence on a single modality that is the text modality. There have been few successful attempts to combine multiple modalities. ImageNet \cite{imagenet2009} and NEIL \cite{chen_iccv13} enriched text with images obtained from Internet search. They used crowd-sourcing and unsupervised learning to get the object labels. These object labels were further extended to object affordances \cite{zhu2014}.

We have seen successful applications of the existing KBs within the modalities they covered, such as IBM Watson Jeopardy Challenge \cite{ferrucci2012a}. However, the existing KBs are human centric and do not directly apply to robotics. The robots need finer details about the physical world, e.g., how to manipulate objects, how to move in an environment, etc. In \robobrain{} we combine knowledge from the Internet sources with finer details about the physical world, from \robobrain{} project partners, to get an overall rich graph representation.

\smallskip

% !TEX root = robobrain.tex

\noindent
\textbf{Robot Learning.}
For robots to operate autonomously they should perceive their environments, plan paths, manipulate objects and interact with humans. We describe previous work in each of these areas and how \robobrain{} complements them.

\noindent
\emph{Perceiving the environment.} Perception is a key element of many robotic tasks. It has been applied to object labeling~\cite{lai:icra11a, KoppulaIJRR2012,wulenzsaxena2014_hierarchicalrgbdlabeling}, scene understanding~\cite{KitaniECCV2012,guptaECCV14}, robot localization~\cite{McManus-RSS-14,NaseerAAAI14},  path planning~\cite{KatzAR14}, and object affordances~\cite{delaitre2012, KoppulaECCV14}. \robobrain{} stores perception related knowledge in the form of 3D point clouds, grasping features, images and videos. It also connects this knowledge to human understandable concepts from the Internet knowledge sources.

\noindent
\emph{Path planning and manipulation.} Planning algorithms formulate action plans which are used by robots to move around and modify its environment. Planning algorithms have been proposed for the problems of motion planning~\cite{ZuckerCHOMP13, SchulmanRSS13},  task planning~\cite{alami2006toward, BolliniISER12} and symbolic planning~\cite{edelkamp2009optimal, rintanen2012planning}. Some planning applications include robots baking cookies~\cite{BolliniISER12}, folding towels~\cite{Maitin-ShepardICRA10}, assembling furniture~\cite{KnepperICRA13}, and preparing pancakes~\cite{Beetz11}. The previous works have also learned planning parameters using methods such as Inverse Optimal Control~\citep{AbbeelIJRR10,Ratliff06,ZiebartAAAI08,jainsaxena2013_trajectorypreferences}. \robobrain{} stores the planning parameters learned by previous works and allow the robots to query for the parameters.

\iffalse
\noindent
\emph{Path and manipulation planning.} There exist a large class of algorithms which allow
robots to move around and modify the environment. Broadly planning algorithms can be
categorized as motion planning \cite{ZuckerCHOMP13, SchulmanRSS13},
 task planning \cite{alami2006toward, BolliniISER12} and symbolic planning
 \cite{edelkamp2009optimal, rintanen2012planning}.
Bakebot \cite{BolliniISER12}, towel-folding \cite{Maitin-ShepardICRA10}, IkeaBot \cite{KnepperICRA13} and robots preparing pancakes \cite{Beetz11}
are few of the many successful planning applications.
Most planning algorithms abstract out the perception details, however access to
perception and manipulation knowledge can allow robots to plan in dynamic real world environments.
\fi

\noindent
\emph{Interacting with humans.} Human-robot interaction includes collaborative tasks between humans and robots~\cite{NikolaidisISR10,nikolaidis2013human}, generating safe and human-like robot motion~\cite{Mainprice13,LasotaCASE14,GielniakIJRR13,DraganRSS13,CakmakIROS11}, interaction through natural language~\cite{TellexRSS14,misra2014tell}, etc. These applications require joint treatment of perception, manipulation and natural language understanding. \robobrain{} stores different data modalities required by these applications.
\iffalse
\noindent
\emph{Interacting with humans.} Another important aspect is
human-robot interaction. Previous works have focused on various aspects,
such as human-robot collaboration for task completion \cite{NikolaidisISR10, koppula-anticipatoryplanning-iser2014}, generating safe
robot motion near humans \cite{Mainprice13, LasotaCASE14},
 obeying user preferences \cite{CakmakIROS11},
 generating human like and legible motions \cite{GielniakIJRR13, DraganRSS13},
interaction through natural language \cite{TellexRSS14,misra2014tell}, etc.
These applications require joint treatment of perception, manipulation and natural language understanding.

%All these applications require access to perception, manipulation, language understanding, etc.,
%further demonstrating the need for large scale multi-modal data.

%Although, there has been significant success in addressing these problems individually,
%previous works were not able to exploit the common information shared by these problems.
%For example, \todo{rewrite the example below and clarify why they couldnt use the shared information..
%However, these problems share many common information which previous works have not exploited.
%Such as previous works have perceived
%environments with cups, have manipulated with for pouring water and have learned
%how to manipulate it near human when it contains hot coffee. }
%Our RoboBrain knowledge engine enables sharing information across these tasks by allowing
%queries to the graph for information spanning several sub-problems.
\fi

Previous efforts on connecting robots range from creating a common operating system (ROS) for
 robots \cite{Quigley09}
to sharing data acquired by various robots via cloud \cite{RoboEarth, KIVA}.
For example, the RoboEarth \cite{RoboEarth} provides a platform for the robots to store and off-load computation
to the cloud and communicate with other robots; and the KIVA systems \cite{KIVA} use the cloud to coordinate motion for hundreds of mobile platforms. On the other hand, \robobrain{} provides a
knowledge representation layer on top of data storing, sharing and communication.

Open-Ease~\cite{openease} is a related on-going effort towards building a knowledge engine for robots.
Open-Ease and \robobrain{} differ in the way they learn and represent knowledge. In Open-Ease the knowledge is represented as formal statements using pre-defined templates. On the other hand, the knowledge in \robobrain{} is represented as a graph. The nodes of the \robobrain{} graph have no pre-defined templates and they can be any robotic concept like grasping features, trajectory parameters, and visual data. This graph representation allows partner projects to easily integrate their learned concepts in \robobrain{}. The semantic meaning of concepts in the \robobrain{} graph are represented by their connectivity patterns in the graph. %The Open-Ease is similar to the \robobrain{} in that it also collects knowledge through robot interactions.
\iffalse
A closely related on-going effort is Open-Ease \cite{openease}
It focuses on a similar problem of cloud knowledge-base and inference engine for robots.
It is different from \robobrain{} in a way it incorporates knowledge and do inference.
Open-ease requires input to be in the form of a facts written in a formal logic statements following the pre-defined structure and uses an
inference engine designed for the language. On the other hand, \robobrain{} accepts unstructured information as nodes and their relations as edges.
\robobrain{} automatically infers the structure and it also infers the correctness and the latent semantic meaning. Our approach brings flexibility with the cost
of a technical challenges in inference and learning.
\fi

% \input{properties}

% !TEX root = robobrain.tex

% \section{Technical Challenges}
\section{Overview}
\label{overviewPaper}

\robobrain{} is a never ending learning system  that  continuously incorporates
new knowledge from  its partner projects and from different Internet sources.
One of the functions of \robobrain{} is to represent the knowledge from various sources as a graph,
as shown in  Figure~\ref{fig:graph}. The nodes of the graph represent concepts and edges represent the
relations between them. The connectivity of the graph is increased through a set of graph operations
that allow additions, deletions and updates to the graph. As of the date of this submission,
\robobrain{} has successfully
connected knowledge from sources like WordNet, ImageNet, Freebase, OpenCyc,
 parts of Wikipedia and other partner projects. These knowledge sources provide lexical knowledge, grounding of concepts into images and common sense facts about the world.

The knowledge from the partner projects and Internet sources can sometimes be erroneous. \robobrain{}
handles inaccuracies in  knowledge by maintaining beliefs over the correctness of the concepts and
relations. These beliefs depend on how much \robobrain{} trusts a given source of knowledge, and also the
feedback it receives from crowd-sourcing (described below). For every incoming knowledge, \robobrain{}
also makes a sequence of decisions on whether to form new nodes, or edges, or both. Since the
knowledge carries semantic meaning \robobrain{} makes many of these decisions based on the
contextual information that it gathers from nearby nodes and edges. For example, \robobrain{} resolves
polysemy using the context associated with nodes. Resolving polysemy is important because a `plant'
could mean a `tree' or an `industrial plant' and merging the nodes together will create errors in the
graph.

\robobrain{} incorporates supervisory signals from humans in the form of crowd-sourcing feedback. This
feedback allows \robobrain{} to update its beliefs over the correctness of the knowledge, and to modify the
graph structure if required. While crowd-sourcing feedback was used in some previous works as
means for data collection (e.g.,~\citep{imagenet2009,Russell08}), in \robobrain{} they serve as supervisory
signals that improve the knowledge engine. \robobrain{} allows user interactions at multiple levels: (i)~
Coarse feedback: these are binary feedback where a user can ``Approve'' or ``Disapprove'' a concept
in \robobrain{} through its online web interface; (ii) Graph feedback: these feedback are elicited on 
\robobrain{} \textit{graph visualizer}, where a user modifies the graph by adding/deleting nodes or edges; (iii)
Robot feedback: these are the physical feedback given by users directly on the robot.

In this paper we discuss different aspects of \robobrain{}, and show how \robobrain{} serves as a knowledge layer for the robots. In order to support knowledge sharing, learning, and crowd-sourcing feedback we develop a large-scale distributed system. We describe the architecture of our  system in Section~\ref{sec:system}. In Section~\ref{sec:raquel} we describe the robot query library, which allow robots to interact with \robobrain{}.  Through experiments we show that robots can use \robobrain{} \textit{as-a-service} and that knowledge sharing through \robobrain{} improves existing robotic applications. We now present a formal definition of our Robot Knowledge Engine and the graph.

% \input{crowdsourcing}

% !TEX root = robobrain.tex

\begin{figure*}[t]
\begin{subfigure}[b]{0.3\textwidth}
\includegraphics[width=\textwidth]{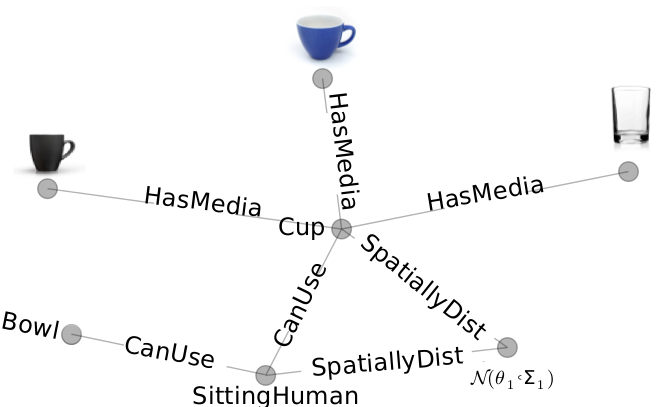}
\caption{original graph}
\end{subfigure}~
\begin{subfigure}[b]{0.3\textwidth}
\includegraphics[width=\textwidth]{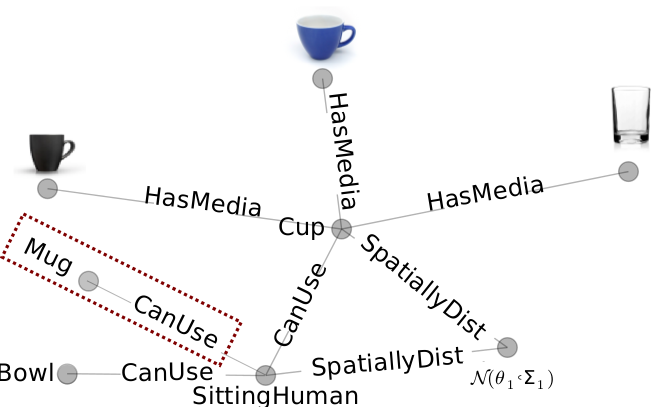}
\caption{feed insertion }
\end{subfigure}~
\begin{subfigure}[b]{0.3\textwidth}
\includegraphics[width=\textwidth]{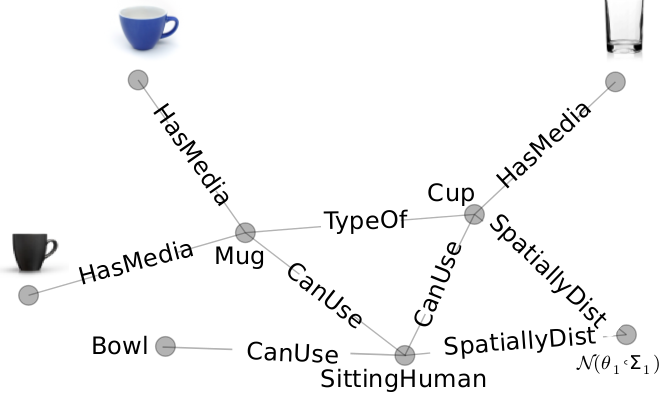}
\caption{after $merge(Mug,Mug^\prime) \rightarrow Mug \circ split(Cup)\rightarrow(Cup,Mug^\prime)$}
\end{subfigure}
\vskip -.05in
\caption{\textbf{Visualization of inserting new information.} We insert `\emph{Sitting human can use a mug}' and \robobrain{} infers the necessary split and merge operations on the graph. In (a) we show the original sub-graph, In (b) information about a \emph{Mug} is seen for the first time and the corresponding node and edge are inserted, In (c) inference algorithm infers that previously connected cup node and  cup images are not valid any more, and it splits the $Cup$ node into two nodes as $Cup$ and $Mug^\prime$ and then merges $Mug^\prime$ and $Mug$ nodes.}
\label{insertgraph}
\vskip -.2in
\end{figure*}

\section{Knowledge Engine: Formal Definition}
\label{sec:graph}

In this section we present the formal definition of \mbox{\robobrain{}}.
\robobrain{} represents knowledge as a directed graph $\mathcal{G}=(V,E)$.
The vertices $V$ of the graph stores concepts that can  be of a variety
of types such as images, text, videos, haptic data, or  learned entities such as affordances, deep learning features, parameters, etc.
The edges $E \subseteq V\times V\times C$ are directed and represents the relations between concepts. Each edge has an edge-type from a set $C$ of possible edge-types.

An edge $(v_1,v_2,\ell)$ is an ordered set of two nodes $v_1$ and $v_2$ and an edge-type  $\ell$.
%connected by an edge type $\ell$ from $v_1$ to $v_2$.
% in which it represent there exist an edge from node $v_1$ to node $v_2$ with type $l$.
Few examples of such edges are: $($StandingHuman, Shoe, \emph{CanUse}$)$,
$($StandingHuman, $\mathcal{N}(\mu,\Sigma)$, \emph{SpatiallyDistributedAs}$)$
and $($Grasping, DeepFeature$23$, \emph{UsesFeature}$)$. %These edges can be considered as (\emph{subject}, \emph{object}, \emph{predicate}) triplets.
We do not impose any constraints on the type of data that nodes can represent. However, we require the edges to be consistent with \robobrain{} edge set $C$. We further associate each node and edge in the graph with \textit{a feature vector representation} and a \textit{belief}. The feature vector representation of nodes and edges depend on their local connections in the graph, and their belief is a scalar probability over the accuracy of the information that the node or an edge represents. Tables~\ref{tbl:vertices} and~\ref{tbl:edges} show few examples of nodes and edge-types. A snapshot of the graph is shown in  Figure~\ref{fig:graph}.

\subsection{Creating the Graph}
Graph creation consists of never ending cycle of two stages namely, knowledge acquisition and inference. Within the knowledge acquisition stage, we collect data from various sources and during the inference stage we apply statistical techniques to update the graph structure based on the aggregated data. We explain these two stages below.

\smallskip
\noindent{\textbf{Knowledge acquisition:}} \robobrain{} accepts new information in the form of set of edges, which we call a \textit{feed}. A \textit{feed} can either be from an automated algorithm crawling the Internet sources or from one of \robobrain{}'s partner projects.
We add a new \textit{feed} to the existing graph through a sequence of union operations performed on the  graph. These union operations are then followed by an inference algorithm.
More specifically, given a new \emph{feed} consisting of a set of $N$ edges \{$(v^1_1,v^1_2,\ell^1) \ldots (v^N_1,v^N_2,\ell^N)$\}, and the existing graph $G = (V,E)$. The graph union operations give a graph $G^\prime=(V^\prime,E^\prime)$ as follows:
\begin{equation}
\label{eq:update}
\begin{aligned}
V^\prime &= v^1_1 \cup v^1_2 \cup \ldots \cup v^N_1 \cup v^N_2 \cup V \\
E^\prime &=  (v^1_1,v^1_2,\ell^1) \cup \ldots \cup (v^N_1,v^N_2,\ell^N) \cup E
\end{aligned}
\end{equation}

%\noindent
%After the knowledge is acquired, we update the graph in the next step.

\smallskip
\noindent{\textbf{Inference on the Graph:}}
After adding the \textit{feed} to the graph using equation~\eqref{eq:update}, we perform inference to update the graph based on this new knowledge. The inference outputs a sequence of graph operations which are then performed on the graph. These graph operations modify the graph by adding new nodes or edges to the graph, deleting nodes or edges from the graph, merging or splitting nodes, etc.

We mention two graph operations here: \emph{split} and \emph{merge}. The split operation is defined as splitting a node into a set of two nodes. The edges having end points in the split node are connected to one of the resultant nodes using the inference algorithm. A merge operation is defined as merging two nodes into a single node, while updating the edges connected to the merged nodes. An example of such an update is shown in Figure \ref{insertgraph}. When a new information \emph{``sitting human can use a mug"} is added to the graph, it causes the \textit{split} of the \emph{Cup} node into two nodes: a \emph{Cup} and a \emph{Mug} node. These two are then connected by an edge-type \textit{TypeOf}. The graph update can be expressed through the following equation:
$$G^\star = split_{v_{s_1}} \circ merge_{v_{m_1},v_{m_2}} \circ \ldots \circ split_{v_{s_M}} \circ G^\prime$$

In the above equation $G^\star$ is the graph obtained after the inference. The goal of the inference steps is to modify the graph $G^\prime$ in a way that best explains the physical world. However, the graph that captures the real physical world is a  \textit{latent graph}, i.e., it is not directly observable.
For  example, the latent information that \emph{``coffee is typically in a container"} is partially observed through
many edges between the \emph{coffee} node and the nodes with \emph{container} images. Our graph construction can also be explained in a generative setting of having a latent graph with all the knowledge about physical word, and we only observe  noisy measurements in form of \textit{feeds}. In this paper, we abstract the algorithmic details of inference and focus on the overall ideas involved in \robobrain{}, its architecture, and its application to robotics.
%system architecture and presenting the overall ideas in the RKE, and defer the details of inference algorithms and the latent graph to a later journal version.

\begin{table}
\caption{Some examples of different node types in our \robobrain{} graph. For full-list,
please see the code documentation.}
\label{tbl:vertices}
\begin{tabular}{ll}
Word & an English word represented as an ASCII string\\
DeepFeature & feature function trained with a Deep Neural Network\\
Image & 2D RGB Image\\
PointCloud & 3D point cloud\\
Heatmap & heatmap parameter vector\\
\end{tabular}
\end{table}
\begin{table}
\caption{Some examples of different edge types in our \robobrain{} graph. For full-list,
please see the code documentation.}
\label{tbl:edges}
\begin{tabular}{ll}
IsTypeOf & human \emph{IsTypeOf} a mammal \\
HasAppearance & floor \emph{HasAppearance} as follows (this image) \\
CanPerformAction & human \emph{CanPerformAction} cutting \\
SpatiallyDistributedAs & location of human is \emph{SpatiallyDistributedAs} \\
IsHolonym & tree  \emph{IsHolonym} of leaf
\end{tabular}
\vskip -.2in
\end{table}

% !TEX root = robobrain.tex
\vspace*{2\sectionReduceTop}
\section{System Architecture}
\vspace*{4\sectionReduceBot}
\label{sec:system}
\begin{figure}[t]
\vskip -.15in
\centering
\includegraphics[width=\linewidth]{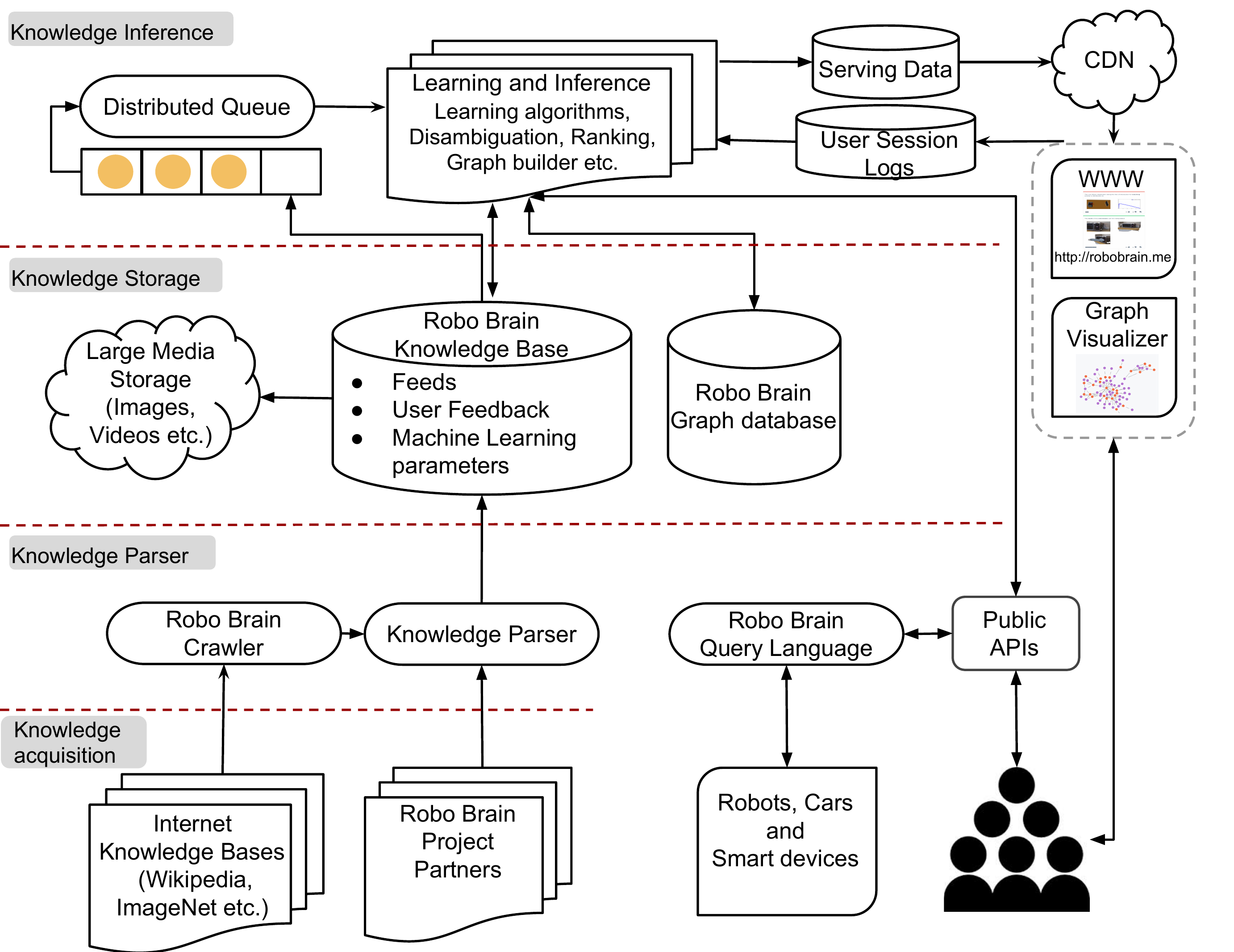}
%\vskip -.1in
\vspace*{3\captionReduceBot}
\caption{\textbf{\robobrain{} system architecture.} It consists of four
interconnected knowledge layers and supports various mechanisms for users and
robots to interact with \robobrain{}.}
\vspace*{2.4\captionReduceBot}
\label{fig:system}
\vskip -.1in
\end{figure}

% \todo {This is just a first pass to set the basic coverage of the section}

We now describe the system architecture of \robobrain{}, shown in Figure~\ref{fig:system}. The system
consists of four interconnected layers: (a) knowledge acquisition, (b) knowledge
parser, (c) knowledge storage,  and (d) knowledge inference. The principle
behind our design is to efficiently process large amount of unstructured
multi-modal knowledge and represent it using the structured \mbox{\robobrain{}} graph.
In addition, our design also supports various mechanisms for users and robots to interact with \robobrain{}. %Through these interactions users retrieve knowledge and also improve \robobrain{} through feedback. 
Below we discuss each of the components.
%and (e) interaction mechanisms.

% It also comprises interactions with robots and users via physical and online interactions.

% many layers and most of the components are restful (Need to write more).

\textit{Knowledge acquisition} layer is the interface between \robobrain{} and the
different sources of multi-modal data. Through this layer \robobrain{} gets access to new information which the other layers process.  \robobrain{} primarily collects knowledge through its partner projects and by crawling the existing knowledge bases such as Freebase, ImageNet and WordNet, etc.,  as well as unstructured sources such as  Wikipedia.

\textit{Knowledge parser} layer of \robobrain{} processes the data acquired by the acquisition layer and converts it to a consistent format for the storage layer. It also marks the incoming data with appropriate meta- data such as  timestamps, source version number etc., for scheduling and managing future data processing. Moreover, since the knowledge bases might change with time, it adds a back pointer to the original source.

\textit{Knowledge storage} layer of \robobrain{} is responsible for storing different representations of the data. In particular it consists of a NoSQL document storage database cluster -- \robobrain{} Knowledge Base (\robobrain{}-KB) -- to store ``feeds'' parsed by the knowledge parser, crowd-sourcing feedback from users, and parameters of different machine learning algorithms provided by \robobrain{} project partners. \robobrain{}-KB offloads large media content such as images, videos and 3D point clouds to a distributed object storage system built using Amazon Simple Storage Service (S3). The real power of \robobrain{} comes through its graph database (\robobrain{}-GD) which stores the structured knowledge. The data from \robobrain{}-KB is refined through multiple learning algorithms and its graph representation is stored in \robobrain{}-GD. The purpose behind this design is to keep \robobrain{}-KB as the \robobrain{}'s single source of truth (SSOT). SSOT centric design allows us to re-build \robobrain{}-GD in case of failures or malicious knowledge sources.

\textit{Knowledge inference} layer contains the key processing and machine learning components of \robobrain{}. All the new and recently updated feeds go through a persistent replicated distributed queuing system (Amazon SQS), which are then consumed by some of our machine learning plugins (inference algorithm, graph builder, etc.) and populates the graph database. These plugins along with  other learning algorithms (operating on the entire graph) constitute our learning and inference framework. 

\robobrain{} supports various \textit{interaction mechanisms} to enable robots and users to communicate with the knowledge engine.
We develop a Robot Query Library as a primary method for robots to interact with \robobrain{}. We also make available a set of public APIs to allow information to be presented on the WWW for online learning mechanisms (eg., crowd-sourcing). \robobrain{} serves all its data using a commercial content delivery network (CDN) to reduce the end user latency. 

% !TEX root = robobrain.tex
\section{Robot Query Library (RQL)}
\label{sec:raquel}
In this section we present the RQL query language,
through which the robots use \robobrain{} for various robotic applications.
The RQL provides a rich set of \textit{retrieval functions} and \textit{programming constructs} to
perform complex traversals on the \robobrain{} graph. An example of such a query is finding the possible ways
for humans to use a cup. This query requires traversing paths from the human node to the cup node in
the \robobrain{} graph.

The RQL allows expressing both the \textit{pattern} of sub-graphs to match and the
\textit{operations} to perform on the retrieved information. An example of such an operation is
\textit{ranking} the paths from the human to the cup node in the order of relevance. The RQL admits following two types of functions: (i) graph retrieval
functions; and (ii) programming construct functions.

\subsection{Graph retrieval function}
The graph retrieval function is used to find sub-graphs matching a given \textit{template} of the form: $$\text{Template: }(u)\rightarrow [e] \rightarrow (v)$$
In the template above, the variables $u$ and $v$ are nodes in the graph and the variable $e$ is a directed edge from $u$ to $v$. We represent the graph retrieval function with the keyword ${\tt fetch}$ and the corresponding RQL query takes the following form:
$${\tt fetch}(\text{Template})$$
The above RQL query finds the sub-graphs matching the template. It instantiates the variables in the template to match the sub-graph and returns the list of instantiated variables. We now give a few use cases of the retrieval function for \robobrain{}.

\begin{example}
The RQL query to retrieve all the objects that a human can use
\begin{align*}
\text{Template: }&{\tt(\{name:`Human'\})\rightarrow [`CanUse'] \rightarrow (v)}\\
\text{Query: }&{\tt fetch}(\text{Template})
\end{align*}
The above query returns a list of nodes that are connected to the node with name ${\tt{Human}}$ and with an edge of type ${\tt{CanUse}}$.% We  now give a query example for retrieving nodes that not directly connected.
\end{example}
Using the RQL  we can also express several \textit{operations} to perform on the retrieved results.
The operations can be of type   ${\tt SortBy}$, ${\tt Len}$, ${\tt Belief}$ and ${\tt ArgMax}$. We now
explain some of these operations with an example.
\begin{example}
The RQL query to retrieve and sort all possible paths from the ${\tt{Human}}$ node to the ${\tt{Cup}}$ node.
%\noindent \resizebox{\linewidth}{!}{
%\begin{minipage}{\linewidth}
\vskip -0.13in
{\small
\begin{align*}
 &{\tt paths }:= \;{\tt fetch (\{name:`Human'\})\rightarrow [r *] \rightarrow (\{name:`Cup'\})}\\
& {\tt SortBy(\lambda P \rightarrow Belief \, P) \, paths} \\
\end{align*}
%\end{minipage}
}\vskip -0.15in

In the example above, we first define a function ${\tt paths}$ which returns all  the paths from the node ${\tt Human }$ to the node ${\tt Cup }$ in the form of a list. The ${\tt SortBy}$ query first runs the ${\tt paths}$ function and then sorts, in decreasing order, all paths in the returned list using their beliefs.
\end{example}

\subsection{Programming construct functions}
The programming construct functions serve to process the sub-graphs retrieved by the graph retrieval function ${\tt fetch}$. In order to define these functions we make use of functional programming constructs like ${\tt map}$, ${\tt filter}$ and ${\tt find}$. We now explain the use of some of these constructs in RQL.

\begin{example}
The RQL query to retrieve affordances of all the objects that ${\tt{Human}}$ can use.
\vskip -0.13in
%\noindent \resizebox{\linewidth}{!}{
%\begin{minipage}{\linewidth}
{\small
\begin{align*}
&  {\tt objects }:=\; {\tt fetch (\{name:`Human'\})\rightarrow [`CanUse'] \rightarrow (v)} \\
&  {\tt affordances \;n } := \;{\tt fetch (\{name:n\})\rightarrow [`Has Affordance'] \rightarrow (v)} \\
&  {\tt map(\lambda u \rightarrow affordances \, u) \,  objects} \\
\end{align*}
%\end{minipage}
}\vskip -0.15in

\noindent In this example, we illustrate the use of ${\tt map}$ construct. The ${\tt map}$ takes as input a function and a list, and then applies the function to every element of the list. More specifically, in the example above, the function ${\tt objects }$ retrieves the list of objects that the human can use. The ${\tt affordances }$ function takes as input an object and returns its affordances. In the last RQL query, the ${\tt map}$  applies the function ${\tt affordances}$ to the list returned by the function ${\tt objects}$.
\end{example}
We now conclude this section with an expressive RQL query for retrieving \textit{joint parameters} shared among nodes. Parameters are one of the many concepts we store in \robobrain{} and they represent learned knowledge about nodes. The algorithms use joint parameters to relate multiple concepts and here we show how to retrieve joint parameters shared by multiple nodes. In the example below, we describe the queries for parameter of a single node and parameter shared by two nodes.

\begin{example}
\label{ex:joint}
The RQL query to retrieve the joint parameters shared between a set of nodes.
\vskip -0.13in
%\noindent \resizebox{\linewidth}{!}{
%\begin{minipage}{\linewidth}
{\small
\begin{align}
&{\tt {parents} \,\, n := fetch \,\, (v)\rightarrow [`HasParameters']\rightarrow  (\{handle:n\}) }\nonumber \\
& {\tt {parameters} \,\, n :=  fetch \,\, (\{name:n\})\rightarrow [`HasParameters']\rightarrow }  (v)  \nonumber  \\
& {\tt {ind\_parameters}\;\; a := }  \cr
&{\tt \hspace*{2cm} filter (\lambda u \rightarrow\! {Len \; parents}\; u = 1)     {parameters}\, a \nonumber } \\
& {\tt {joint\_parameters} \,\, a_1 \,\, a_2\,\, := } \cr
& {\tt \hspace*{2cm} filter (\lambda u \rightarrow {Len\; parents}\,\, u = 2\,\, and } \cr
& {\tt \hspace*{2cm}  u \,\, in \,\, {parameters}\,\, a_2)    \,\,{parameters}\,\, a_1 \nonumber}
\end{align}
%\end{minipage}
}\vskip -0.01in
The query above uses the ${\tt filter}$ construct function and ${\tt Len}$ operation. The ${\tt filter}$ takes as input a list and a check condition, and returns only those items from the list that satisfies the input condition. The ${\tt Len}$ takes as input a list and returns the number of items in the list.
In the query above, we first define a function ${\tt parents}$ which for a given input node returns its parent nodes. Then we define a function ${\tt parameters}$ which for a given input node returns its parameters. The third and the fourth queries are functions accepting one and two input nodes, respectively, and return the (joint) parameters that share an edge with every input node and not with any other node.
\end{example}

% !TEX root = robobrain.tex
\vspace*{\sectionReduceTop}
\section{Applications}
\vspace*{3\sectionReduceBot}
\label{sec:applications}
In this section we first show how \robobrain{} can be used \textit{as-a-service} by the robots for several robotics problems. Specifically, we explain the usage of \robobrain{} in anticipating human activities, grounding of natural language sentences, and path planning. We then show how \robobrain{} can help robotics projects by sharing knowledge within the projects and throughout the Internet.

\vspace*{\subsectionReduceTop}
\subsection{\robobrain{} \textit{as-a-service}}
\vspace*{3\subsectionReduceBot}
Our goal with providing \robobrain{} \textit{as-a-service} is to allow robots to use the representations learned by different partner projects. 
% For this purpose the 
% RKE stores the representations learned by different projects, allows queries through RQL. 
This 
% feature of the RKE 
allows \robobrain{} to effortlessly address many robotics applications. In the following we demonstrate \robobrain{} as-a-service feature for three robotics applications that deal with different data modalities of perception, natural language and trajectories.

% !TEX root = robobrain.tex
\begin{figure}
\centering
\includegraphics[width=\linewidth]{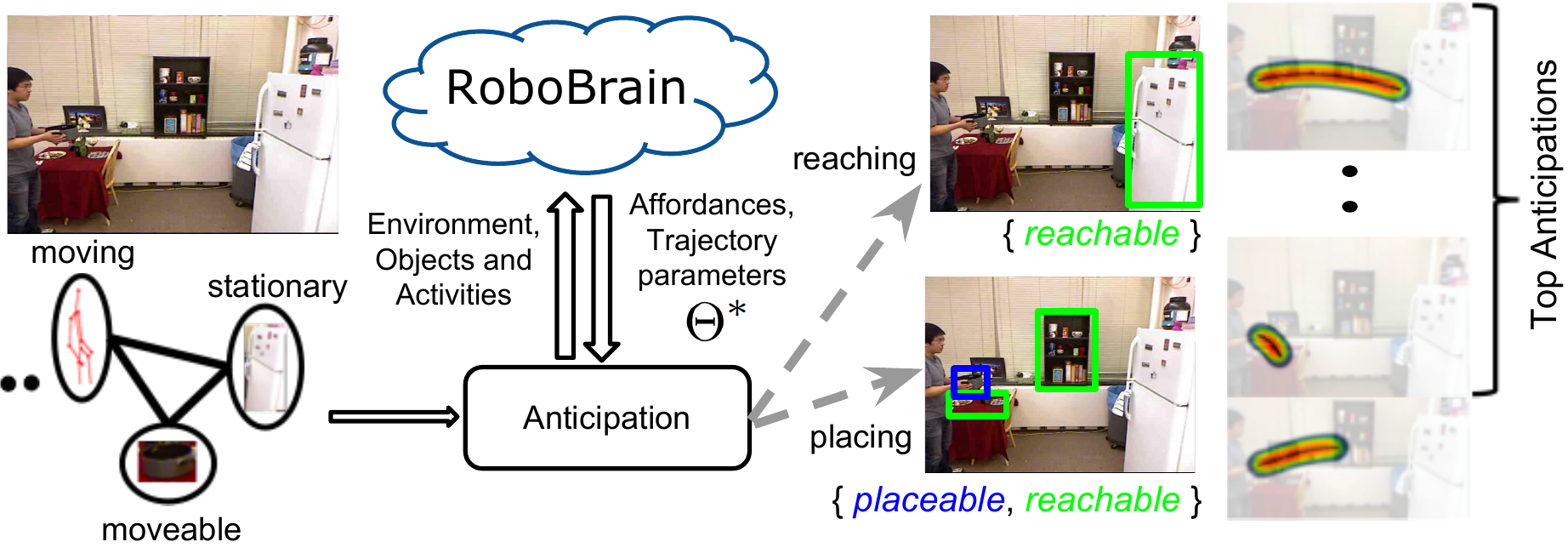}
\vspace*{2\captionReduceBot}
\caption{\textbf{\robobrain{} for anticipating human activities.} Robot using anticipation algorithm of \citet{KoppulaRSS2013} queries \robobrain{}, for the activity, affordance and trajectory parameters in order to generate and rank the possible future activities in a given environment.}
\vspace*{3\captionReduceBot}
\label{Fig:anticipation}
\end{figure}

\subsubsection{Anticipating human actions}
\label{sec:anticipation}
The assistive robots working with humans should be able to understand human activities and also anticipate the future actions that the human can perform. In order to anticipate, the robot should reason over the action possibilities in the environment, i.e., \textit{object affordances}, and how the actions can be performed, i.e., \textit{trajectories}. Several works in robotics have addressed the problem of anticipation~\citep{KitaniECCV2012,KoppulaRSS2013,Kuderer-RSS-12}.

We now show how robots can query \robobrain{} and use the previous work by
Koppula et al.~\citep{KoppulaRSS2013} for anticipating human actions.  
In order to anticipate the future human actions, the authors~\citep{KoppulaRSS2013} learn parameters using their anticipatory algorithm, and using the learned parameters they anticipate the most likely \textit{future} object affordances and human trajectories. \robobrain{} serves anticipation \textit{as-a-service} by storing those learned parameters, object affordances and trajectories  as concepts in its graph. Figure~\ref{Fig:anticipation} illustrates a robot retrieving relevant information for anticipation. The robot first uses the following queries to retrieve the possible trajectories of an object:
\vskip -.15in
%\noindent \resizebox{\linewidth}{!}{
%\begin{minipage}{\linewidth}
{\small
\begin{align}
&{\tt {affordances} \,\, n := fetch \,\,(\{name:n\})\rightarrow   [`HasAffordance'] \rightarrow } \cr
& {\tt \hspace*{2cm} (v\{src:`Affordance'\})   \nonumber } \\
&{\tt {trajectories} \,\, a := fetch \,\,(\{handle:a\})\rightarrow [`HasParameters']\rightarrow } \cr
& {\tt \hspace*{2cm}  (v\{src:`Affordance', type:`Trajectory'\})  \nonumber } \\
& {\tt { trajectory\_parameters} \,\, o := \,\, } \cr
&{\tt \hspace*{2cm} map (\lambda a \rightarrow {trajectories}\,\, a)  \,\, {affordances}\,\, o } \nonumber
\end{align}
%  \end{minipage}
}\vskip -.05in

In the queries above, the robot first queries for the affordances of the object and then for each affordance it queries \robobrain{} for the trajectory parameters. Having retrieved all possible trajectories, the robot uses the learned parameters~\citep{KoppulaRSS2013} to anticipate the future human actions. Since the learned parameters are also stored in the \robobrain{} graph, the robot retrieves them using the following RQL queries:
\vskip -.15in
{\small
\begin{align*}
&{\tt {parents} \,\, n := fetch \,\, (v)\rightarrow [`HasParameters']\rightarrow  (\{handle:n\}) }\nonumber \\
& {\tt {parameters} \,\, n :=  fetch \,\, (\{name:n\})\rightarrow [`HasParameters']\rightarrow }\cr
& {\tt \hspace*{2cm}  (v\{src:`Activity'\})   \nonumber } \\
& {\tt find\_parameters\, a := }\\
& \hspace*{2cm}{\tt filter (\lambda u \rightarrow\! {Len\; parents}\, u = 1)     {parameters}\, a \nonumber } \\
& {\tt {joint\_parameters} \,\, a_1 \,\, a_2\,\, := filter (\lambda u \rightarrow {Len\; parents}\,\, u = 2\,\, } \cr
& {\tt \hspace*{2cm}  and\,\, u \,\, in \,\, {parameters}\,\, a_2)    \,\,{parameters}\,\, a_1 \nonumber}
\end{align*}
}\vskip -.05in
The queries above retrieves both independent and joint parameters for anticipating the object affordances and human activities. Detailed explanation of the query is given in Example \ref{ex:joint} of Section \ref{sec:raquel}

% !TEX root = robobrain.tex

\subsubsection{Grounding natural language} The problem of grounding a natural language instruction in an environment requires the robot to formulate an action sequence  that accomplish the semantics of the  instruction~\citep{tellex2011understanding,misra2014tell,guadarrama2013grounding, MatuszekISER2012}. In order to do this, the robot needs a variety of information. Starting with finding action verbs and objects in the instruction, the robot has  to discover those objects and their affordances in the environment.

% \robobrain{} provides natural language grounding \textit{as-a-service}.

We now show the previous work by Misra et al.~\cite{misra2014tell} using \robobrain{} \textit{as-a-service} in their algorithm. In order to ground a natural language instruction the robot has to check for the satisfiability of the actions it generates in the given environment. For example, an action which pours water on a book should be deemed unsatisfiable. In the previous work~\cite{misra2014tell}, the authors manually define many pre-conditions to check the satisfiability of actions. For example, they define manually that a \textit{syrup bottle} is \textit{squeezable}. Such satisfiability  depends on the object's affordances in the given environment, which can be retrieved from \robobrain{}.

\begin{figure}[t]
\centering
\includegraphics[width=1\linewidth]{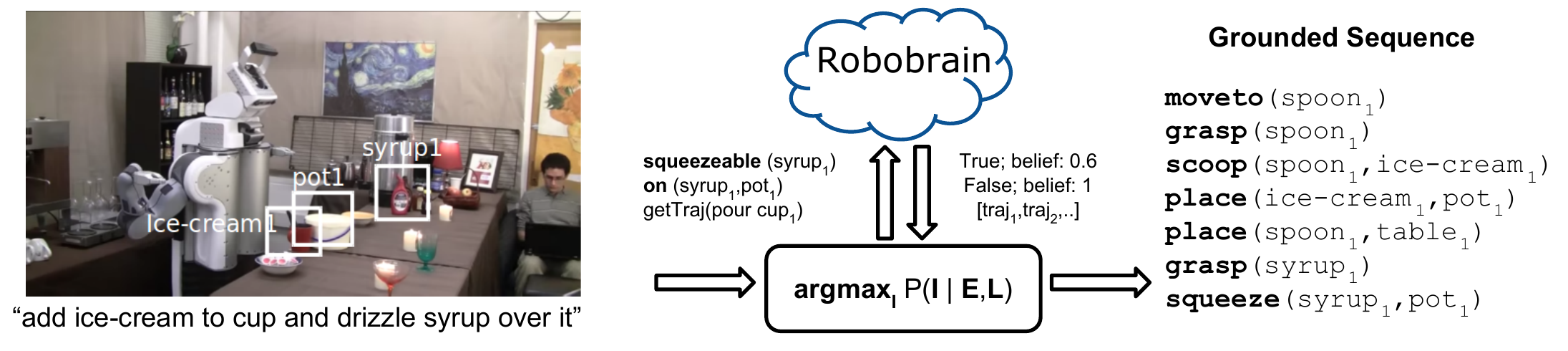}
\vskip -.05in
\caption{\textbf{Grounding natural language sentence.} The robot grounds natural language by using the algorithm by Misra et al.~\cite{misra2014tell} and querying \robobrain{} to check for the satisfiability of actions.}
%\cite{misra2014tell} requires several  pieces of knowledge. Robot should know the appearances and affordances of objects, as well as the possible actions related to each object and it should know the trajectories and manipulation features of those actions. These are all  translated as queries to the RKE graph.}
\vspace*{\captionReduceBot}
\label{Fig:languagegrounding}
\vskip -.05in
\end{figure}

Figure~\ref{Fig:languagegrounding} illustrates a robot querying \robobrain{} to check the satisfiability of actions that it can perform in the given environment. Below is the RQL query for retrieving the satisfiability of  \textit{squeezable} action:
%\noindent \resizebox{\linewidth}{!}{
%\begin{minipage}{\linewidth}
\vskip -.15in
{\small
\begin{align*}
&{\tt {squeezable \,\ syrup}\,\,  := \,\,Len \,\,fetch \,\,(u\{name:`syrup'\})\rightarrow }\\
&{\tt \hspace*{0.1cm} [`HasAffordance']\rightarrow (v\{name:`squeezable'\})\,\,> 0}
\end{align*}
%  \end{minipage}
}\vskip -.01in

% !TEX root = robobrain.tex
\subsubsection{Path planning using \robobrain{}}
\label{sec:applicationplanit}

One key problem robots face in performing tasks in human environments is identifying trajectories desirable to the users. An appropriate trajectory not only needs to be valid from a geometric standpoint (i.e., feasible and obstacle-free), but it also needs to satisfy the user preferences~\citep{jainsaxena2013_trajectorypreferences,Jain14}. For example,  a robot should move sharp objects such as knife strictly away from nearby humans~\cite{jain_contextdrivenpathplanning_2013}. Such preferences are commonly represented as cost functions which jointly model the environment, the task, and trajectories.
%This joint modeling is expensive in terms of resource requirement (e.g., robots) and data collection, and
Typically research groups have independently learned different cost functions~\citep{jainsaxena2013_trajectorypreferences,Kuderer-RSS-12,KitaniECCV2012}, which are not shared across the research groups. Here we show \robobrain{} \textit{as-a-service}  for a robot to store and retrieve the planning parameters.
%We now show how the previous work by Jain et al.~\citep{Jain14} use the RKE for retrieving the trajectory parameters for path planning. We demonstrate this on the example of moving an egg carton from the previous work~\citep{Jain14}.

\begin{figure}[t]
\centering
\includegraphics[width=\linewidth]{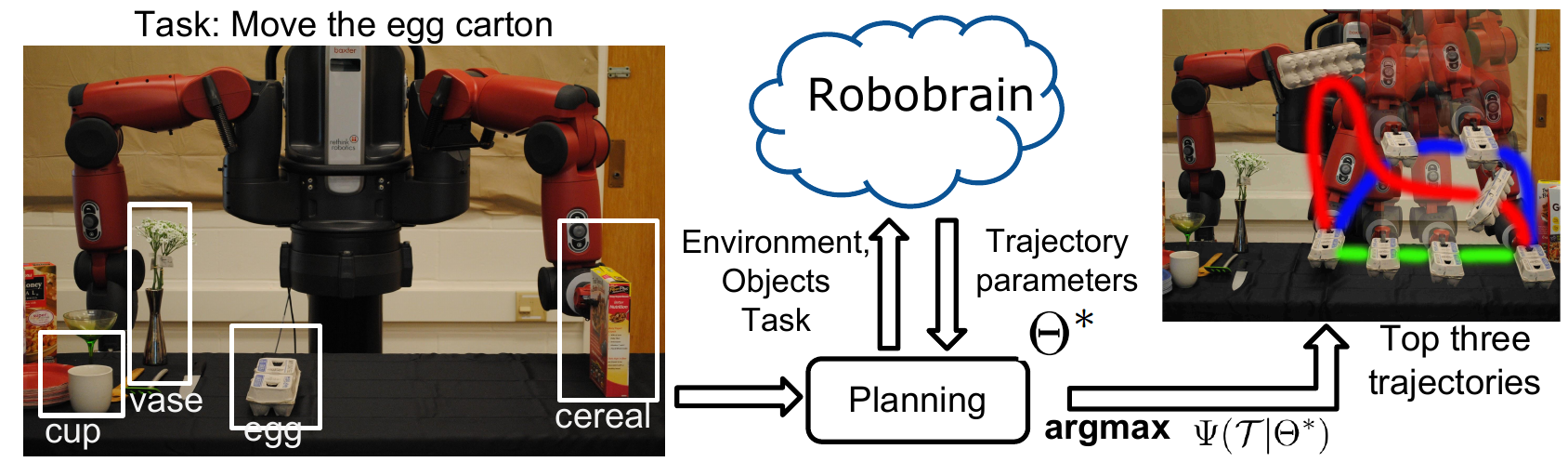}
\caption{\textbf{\robobrain{} for planning trajectory.} The robot queries \robobrain{} for the trajectory parameters (learned by Jain et al.~\citep{jainsaxena2013_trajectorypreferences})  to plan paths for the fragile objects like an egg carton. }
\vspace*{2.5\captionReduceBot}
\label{fig:planning}
\vskip -.1in
\end{figure}

In Figure~\ref{fig:planning} we illustrate the robot planning for an egg carton by querying \robobrain{}.
Since eggs are \textit{fragile}, users  prefer to move them slowly and close to the surface of the table.  In order to complete the task, the robot queries \robobrain{} and retrieves the attributes of the egg carton and also the trajectory parameters learned in the previous work by Jain et al.~\citep{Jain14}. Using the retrieved attributes and the parameters, the robot samples trajectories and executes the top-ranked trajectory.
%need labels of all objects in the environment. In this example we use the
%object labels provided by the authors~\cite{Jain14}. After the object labels are obtained, the previous
%work~\citep{Jain14} use the RKE to retrieve attributes of the egg carton and also the trajectory
%parameters.
Below we show the RQL queries.
%\noindent \resizebox{\linewidth}{!}{
%\begin{minipage}{\linewidth}
\vskip -.15in
{\small
\begin{align*}
&{\tt {attributes} \,\, n := fetch \,\,(\{name:n\})\rightarrow
[`HasAttribute'] \rightarrow (v)   \nonumber} \\
&{\tt {trajectories} \,\, a := }\\
&{\tt\hspace*{2cm} fetch \,(\{handle :a\})\rightarrow  [`HasTrajectory'] \rightarrow (v)  \nonumber } \\
&{\tt {trajectory\_parameters} := }\\
&{\tt \hspace*{2cm} map (\lambda a \rightarrow
{trajectories}\,\, a)  \,\, {attributes}\,\, `egg' } \nonumber
\end{align*}
%  \end{minipage}
}\vskip -.1in

% !TEX root = robobrain.tex

\subsection{\robobrain{} for sharing knowledge}
\robobrain{} allows sharing the knowledge learned by different research groups as well as knowledge obtained from various internet sources. In this section we show with experiments how sharing knowledge improves existing robotic applications:

\subsubsection{Sharing knowledge from the Internet}
In this experiment we show that sharing knowledge from several Internet sources using \robobrain{} improves robotic applications such as path planning.  Knowledge from the Internet sources has been shown to help robots in planing better paths~\citep{beetzIcra2010}, understand natural language~\citep{coyneCosli,tellex2011understanding}, and also recently in object retrieval~\citep{guadarramaRss2014}. However, for certain robotic tasks a single Internet source does not cover many of the real world situations that the robot may encounter. In such situations it is desired to share the information from other sources to get an overall richer representation.
The \robobrain{} graph is designed to acquire and connect information from multiple Internet sources and make it accessible to robots.

In this experiment we build upon work by Jain et al.~\citep{jainsaxena2013_trajectorypreferences} for planning trajectories that follow user preferences. The work relied on object attributes in order to plan desirable trajectories. These attributes convey properties such as whether an object is sharp, heavy, electronic etc. The attributes were manually defined by the authors~\citep{jainsaxena2013_trajectorypreferences}. In practice this is very challenging and time-consuming because there are many objects and many attributes for each object.
%Jain et al.~\citep{jainsaxena2013_trajectorypreferences}  manually define attributes used by TPP.
Instead of manually defining the attributes, we can retrieve many of them from the Internet knowledge
sources such as OpenCyc, Wikipedia, etc. However, a single knowledge source might not have
attributes for all objects.
The \robobrain{} graph connects many attributes obtained from multiple Internet sources to their respective objects.
%We crawl multiple internet sources to retrieve many attributes and connect them to their respective objects in the \robobrain{} graph.

\begin{figure}
\vskip -.16in
\centering
\includegraphics[width=\linewidth]{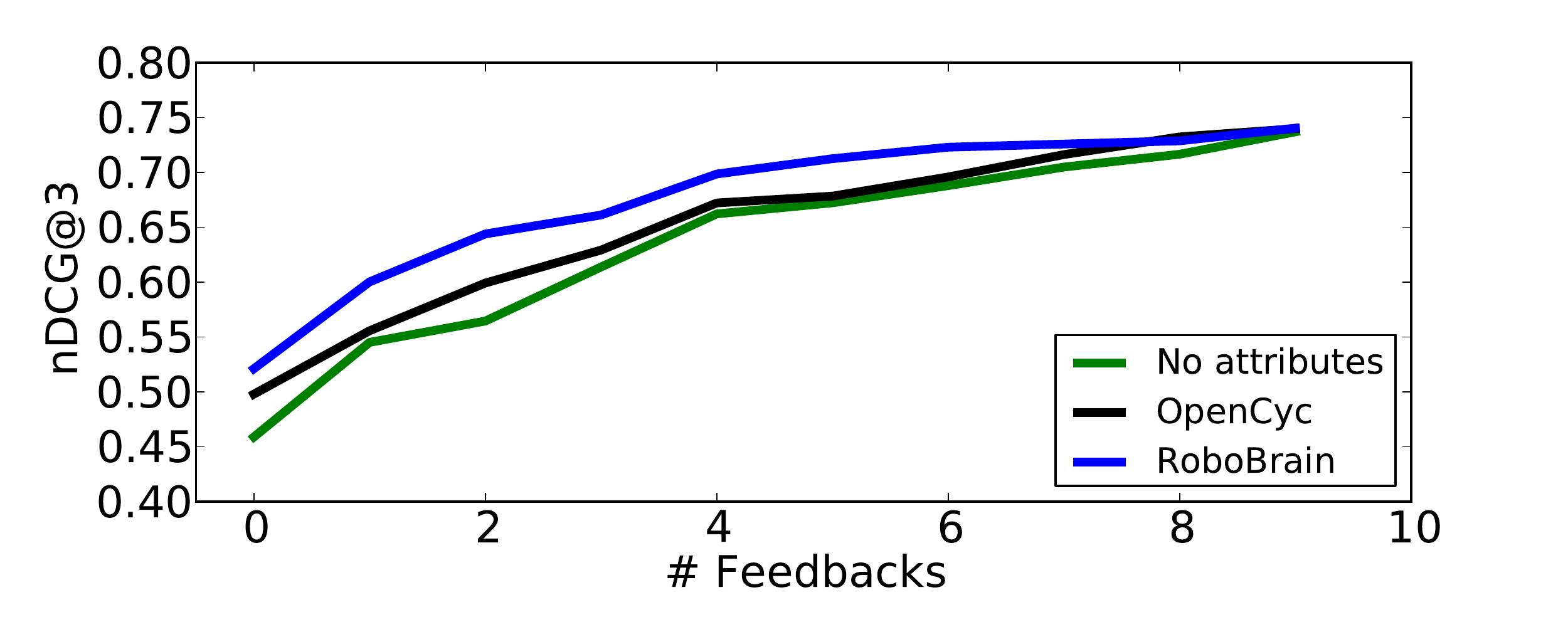}
\vspace*{4\captionReduceTop}
\caption{\textbf{Sharing from Internet sources.} The plot shows performance of  the algorithm by Jain et al.~\citep{jainsaxena2013_trajectorypreferences} for three settings of attributes. This is an online algorithm that learns a good trajectory from the user feedback. The performance is measured using the nDCG metric~\cite{manning2008introduction}, which represents the quality of the ranked list of trajectories. \robobrain{} combines information from multiple sources and hence its richer in attributes as compared to retrieving attributes from OpenCyc alone.}
\vspace*{2\captionReduceBot}
\label{fig:ndcg}
\vskip -.1in
\end{figure}

Figure~\ref{fig:ndcg} illustrates the planning results when the robot does not use any attributes, when it uses attributes from a single source (OpenCyc), and when it use attributes from \robobrain{}. The planning performance is best when using \robobrain{} since it covers more attributes than the OpenCyc alone. Most importantly all these attributes are retrieved from the \robobrain{} graph with a single RQL query as explained in Section \ref{sec:applicationplanit}.

\subsubsection{Sharing learned representations}

New algorithms are commonly proposed for a problem to address the shortcomings of previous methods. These algorithms have their own learned representations. For example,  different representations  have been learned for grounding natural language~\citep{tellex2011understanding,misra2014tell,guadarrama2013grounding, MatuszekISER2012}. However, it is usually hard for practitioners to choose a single representation since there are always inputs where one representation fails but some others work. In this experiment we show that a robot can query \robobrain{} for the best representation while being agnostic to the algorithmic details of the learned representations.

\iffalse
It is common for different research groups to address the same problem by learning new representations. For example,  different representations  have been learned for grounding natural language~\citep{tellex2011understanding,misra2014tell,guadarrama2013grounding, MatuszekISER2012}. It is usually hard for practitioners to choose to a single representation, since there are always inputs where one representation fails but some other works. \todo{Here we show how \robobrain{} chooses between the representations for outputting a likelihood score using beliefs and ranking of the queries.}
\fi

Simulating the above setting, we present an experiment for sharing multiple learned representations on a natural language grounding problem. Here the goal is  to output a sequence of instructions for the robot to follow, given an input natural language command and an environment.
Following the work by Misra et al.~\citep{misra2014tell}, we train a baseline algorithm for the task of \textit{making ramen} (Algorithm A), and train their full algorithm for the task of \textit{making affogato} (Algorithm B). These algorithms assign a  confidence score (i.e., probability) to the output sequence of instructions.  We store these learned representations as concepts in the \robobrain{} graph, along with a prior belief over the correctness of the algorithms. The robot queries \robobrain{} for a representation as follows:
%\resizebox{\linewidth}{!}{
%\begin{minipage}{\linewidth}
\vskip -.15in
{\small
\begin{align*}
& {\tt algParam  :=  fetch (u\{type:'GroundingAlgorithm'\})}\rightarrow \\
& {\tt \hspace*{2cm} [`HasParameters'] \rightarrow (v)}   \\
& {\tt prior \,\, n :=  fetch (\{name:n\})\rightarrow [`HasPriorProb'] \rightarrow (v)}\\
& {\tt groundings \,\, L,E :=   argMaxBy(\lambda(u,v)\rightarrow v)} \\
& {\tt \hspace*{2cm} map(\lambda(u,v) \rightarrow u(L,E,v)*prior\, u) \, \, \, algParam}
\end{align*}
%\end{minipage}
}\vskip -.1in

\medskip
In the ${\tt algParam}$ function, we retrieve all natural language grounding algorithms from the \robobrain{} graph with their parameters. This returns a list in which each element is a tuple of algorithm $u$ and its parameters $v$. The ${\tt prior}$ function retrieves the prior belief over the correctness of an algorithm. In order to ground a given natural language command ${\tt L}$ in environment ${\tt E}$, the ${\tt grounding}$ function evaluates the likelihood score for each algorithm  using their parameters as ${\tt u(L,E,v)}$. It further incorporates the prior belief over the algorithms, and returns the representation with the highest likelihood score. These set of queries corresponds to the following likelihood maximization equation:
\begin{equation*}
\mathcal{I}^*  = \argmax_{\mathcal{I},m'\in \{A,B\}} P(\mathcal{I}|E, L,  w_{m'}^* ,m')P(m')
\end{equation*}
As shown in the Table~\ref{tbl:grounding-results}, choosing a representation by querying the \robobrain{} achieves better performance than the individual algorithms.

\begin{table}
\caption{\robobrain{} allows sharing learned representations. It allows the robot to query \robobrain{} for a representation given an input natural language command. In this table the Algorithm $A$ is a greedy algorithm based on Misra et al.~\cite{misra2014tell}, and Algorithm $B$ is their full model. The $IED$ metric measures the string-edit distance and the $EED$ metric measures the semantic distance between the ground-truth and the inferred output instruction sequences. The metrics are normalized to 100 such that higher numbers are better.}
\vspace*{0.5\captionReduceBot}
\label{tbl:grounding-results}
\centering
\begin{tabular}{l|cc}
\hline
\textbf{Algorithm} & \textbf{IED} & \textbf{EED}\\
\hline
Algorithm A & 31.7 & 16.3\\
Algorithm B & 23.7 & \textbf{27.0}\\
\robobrain{} (A+B) & \textbf{34.2} & 24.2\\
\hline
\end{tabular}
\end{table}

\iffalse
we show how robots can use \robobrain{} to
The existing robotics projects have learned many representation of the world in order to perform various tasks. However, these learned knowledges are typically not shared across related problems. \robobrain{} system stores these learned representations and various applications can query them through RQL. Hence, \robobrain{} helps to share such a knowledge effortlessly. \robobrain{} system supports various robotics related query and we demonstrate them for three applications: (i) anticipating human actions~\citep{KoppulaRSS2013}; (ii) grounding natural language~\citep{misra2014tell}; and (iii) path planning~\citep{jainsaxena2013_trajectorypreferences}.
\fi

% !TEX root = robobrain.tex
\vspace*{2\sectionReduceBot}
\section{Discussion and Conclusion}
\begin{figure}[h!]
  \vspace{-2mm}
 \includegraphics[width=\linewidth, height=1.5in]{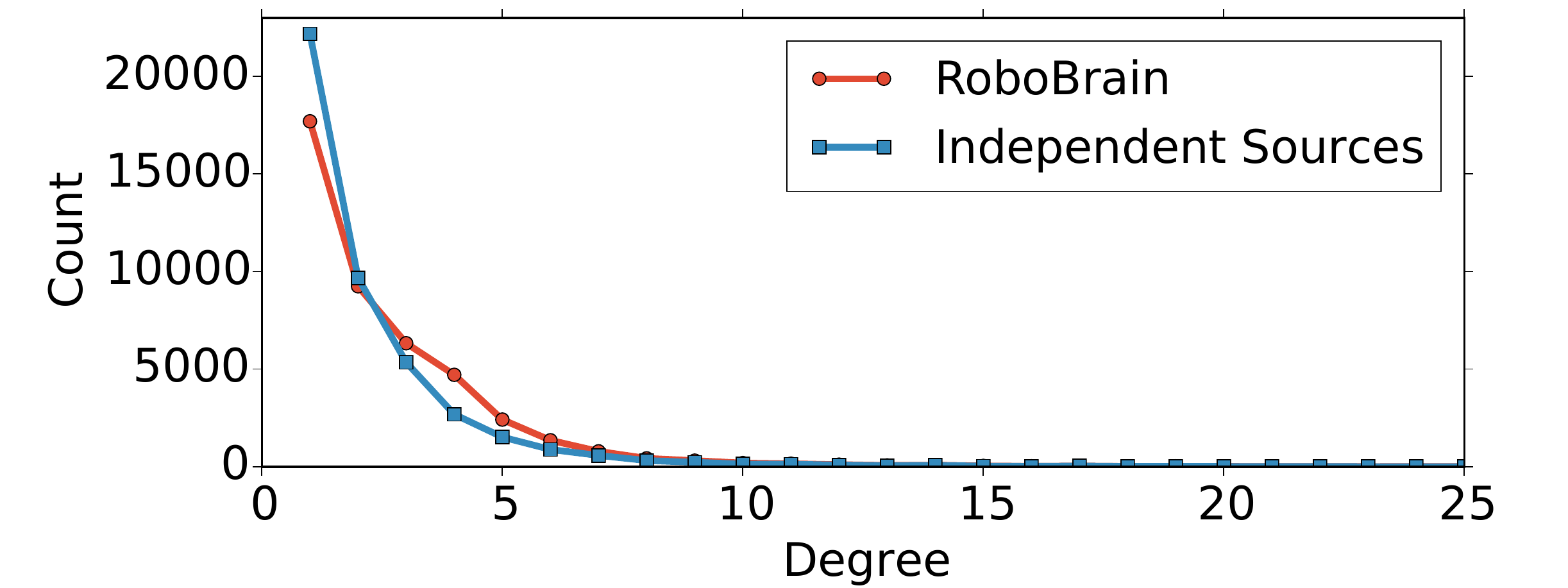}
 \vspace*{3\captionReduceTop}
 \caption{Degree distribution of \robobrain{} and the union of independent knowledge sources.
 For the case of independent sources, we only consider the edges between nodes from the same
 source. \robobrain{} connects different projects successfully: number of nodes with degree 1 and 2 decrease and  nodes with degree 3 and more increase.}
% \vspace*{\captionReduceTop}
\label{degDis}
\vskip -.1in
\end{figure}

\vspace*{2\sectionReduceBot}
The \robobrain{} graph currently has 44347 nodes (concepts) and 98465 edges (relations). The knowledge in the graph is obtained from the Internet sources and through the \robobrain{} project partners. For the success of many robotics application it is important to relate and connect the concepts from these different knowledge sources. In order to empirically evaluate the connectivity of  concepts in \robobrain{}, we plot the degree distribution of the \robobrain{} graph and compare it with the degree distribution of independent knowledge sources (Figure~\ref{degDis}).
The graph of independent knowledge sources is the union of each knowledge source, which have nodes from all the projects and the edges only between the nodes from the same project.
As shown in the Figure~\ref{degDis}, \robobrain{} successfully connects projects and increases the average degree per-node by $0.8$.
The \robobrain{} graph has fifteen thousand nodes with degree one. Most of the nodes with a single degree come from the Internet sources such as Wikipedia  and WordNet. These nodes are not directly related to the physical world and represent abstract concepts like political ideas, categories of art, etc.

In this paper we described different aspects and technical challenges in building \robobrain{} knowledge engine. \robobrain{} represents multiple data modalities from various sources, and connects them to get an overall rich graph representation.  We presented an overview of the \robobrain{} large-scale system architecture and developed the Robot Query Library (RQL) for robots to use \robobrain{}. We illustrated robotics applications of anticipation, natural language grounding, and path planning as simple RQL queries to \robobrain{}. We also showed in experiments that sharing knowledge through \robobrain{} improves existing path planning and natural language grounding algorithms.
 \robobrain{} is an ongoing effort where we are collaborating with different research groups. We are working on improving different aspects such as learning from crowd-sourcing feedback, inference methods over the graph for discovering new relations between concepts, and expanding \robobrain{} to new robotics applications.

\iffalse
\robobrain{} is an ongoing effort, where we are constantly improving different aspects of the work.
% We are improving the system architecture and expanding RQL to support scaling to even larger
% knowledge sources (e.g., millions of videos).
% Furthermore,  w
We have several ongoing research efforts that include achieving
better disambiguation and improving never-ending learning abilities.
More importantly, we are constantly expanding the set of our \robobrain{} research partners.
This will not only improve the abilities of their robots, but also their contribution of knowledge
to \robobrain{} will help other researchers in the robotics community at large.
\fi

\section*{Acknowledgments}
We thank Arzav Jain for his help in building the graph infrastructure for the \robobrain{}.
We thank Bart Selman, Jitendra Malik and Ken Goldberg for their visionary ideas.  We thank
Emin Gun Sirer and Ayush Dubey for help with databases, and thank Silvio Savarese,
Stefanie Tellex, Fei-Fei Li,
and Thorsten Joachims for useful discussions.  We also thank Yun Jiang, Ian Lenz,
 Jaeyong Sung,  and Chenxia Wu for their contributions to the knowledge for \robobrain{}.
 We also thank Michela Meister, Hope Casey-Allen, and Gabriel Kho for their help and involvement
 with the \robobrain{} project.  We are also very thankful to Debarghya Das and Kevin Lee for
 their help with developing the front-end for \robobrain{}.

 This work was supported in part by  Army Research Office (ARO) award W911NF-12-1-0267,  Office of Naval Research (ONR) award N00014-14-1-0156, NSF National Robotics Initiative (NRI) award IIS-1426744, Google Faculty Research award (to Saxena), and Qualcomm research award.
 This was also supported in part by Google PhD Fellowship to Koppula, and by
 Microsoft Faculty Fellowship, NSF CAREER Award and Sloan Fellowship to Saxena.
% (Saxena), and Google Faculty Research Award

%% Use plainnat to work nicely with natbib.

%\newpage

{\small
\bibliographystyle{plainnat}
\bibliography{shortstrings,references}
}

\end{document}